\documentclass{article}

\usepackage{microtype}
\usepackage{graphicx}
\usepackage{subfigure}
\usepackage{booktabs} 

\usepackage{hyperref}

\usepackage[accepted]{icml2025}

\usepackage{amsmath}
\usepackage{amssymb}
\usepackage{mathtools}
\usepackage{amsthm}

\usepackage[capitalize,noabbrev]{cleveref}

\theoremstyle{plain}

\theoremstyle{definition}

\theoremstyle{remark}

\usepackage[textsize=tiny]{todonotes}

\newcommand{\COMMENTALGO}[1]{\textcolor{olive}{\texttt{\small \% #1}}}

\newcommand{\mosa}{\textsc{MoSA}}
\newcommand{\ourmethod}{{\textsc{MoSA}}}

\usepackage{subcaption}

\usepackage{tcolorbox}

\icmltitlerunning{Multi-LLM Collaborative Search for Complex Problem Solving}

\begin{document}

\twocolumn[
\icmltitle{Multi-LLM Collaborative Search for Complex Problem Solving}



\icmlsetsymbol{equal}{*}

\begin{icmlauthorlist}
\icmlauthor{Sen Yang}{cuhk}
\icmlauthor{Yafu Li}{shai}
\icmlauthor{Wai Lam}{cuhk}
\icmlauthor{Yu Cheng}{cuhk,shai}
\end{icmlauthorlist}

\icmlaffiliation{cuhk}{The Chinese University of Hong Kong}

\icmlaffiliation{shai}{Shanghai AI Laboratory}



\icmlkeywords{Machine Learning, ICML}

\vskip 0.3in
]



\printAffiliationsAndNotice{}  

\begin{abstract}
Large language models (LLMs) often struggle with complex reasoning tasks due to their limitations in addressing the vast reasoning space and inherent ambiguities of natural language. 
We propose the Mixture-of-Search-Agents (\mosa{}) paradigm, a novel approach leveraging the collective expertise of multiple LLMs to enhance search-based reasoning. 
\mosa{} integrates diverse reasoning pathways by combining independent exploration with iterative refinement among LLMs, mitigating the limitations of single-model approaches. 
Using Monte Carlo Tree Search (MCTS) as a backbone, \mosa{} enables multiple agents to propose and aggregate reasoning steps, resulting in improved accuracy. 
Our comprehensive evaluation across four reasoning benchmarks demonstrates \mosa{}'s consistent performance improvements over single-agent and other multi-agent baselines, particularly in complex mathematical and commonsense reasoning tasks.
\end{abstract}


\section{Introduction}
\label{sec:intro}

Large language models (LLMs) face challenges with complex reasoning, even when augmented with linearized reasoning chains (e.g., Chain-of-Thought), due to the vast reasoning space inherent in the complexity and ambiguity of natural languages.
A promising approach is step-wise search-based reasoning, which decomposes the reasoning problem into a traversal over a directed graph, where nodes and edges represent individual reasoning sub-steps distributed across the expansive reasoning space.
Related methods have applied various search algorithms to LLMs, such as breadth-first search (BFS), depth-first search (DFS)~\cite{yao2024tree,besta2024graph}, and best-first search~\cite{hao-etal-2023-reasoning,zhang2024accessinggpt4levelmathematical,qi2024mutual}.

\begin{figure}
    \centering
    \includegraphics[width=0.76\linewidth]{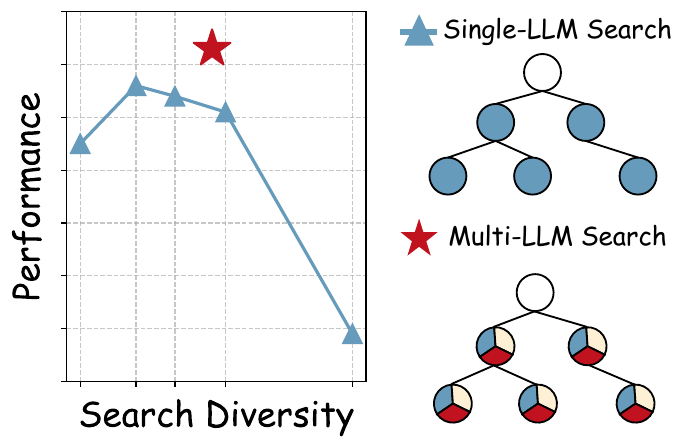}
    \caption{
    Reasoning performance on MATH-500 against search trajectory diversity. 
    While the diversity of single-LLM search varies with different sampling temperatures, the multi-LLM search consistently achieves superior performance. 
    More details are provided in $\S$~\ref{sec:analysis:diversity}.
    }
    \label{fig:intro}
    \vskip -0.2in
\end{figure}

A successful search trial is featured with diverse yet effective explorations~\cite{hao-etal-2023-reasoning,yao2024tree}.
A straightforward method to enhance diversity involves increasing the temperature, thereby making the probability distribution more uniform.
This is typically combined with top-$k$ and top-$p$ sampling to balance diversity and quality.
However, as shown in Figure~\ref{fig:intro}, despite these sampling techniques, achieving a balance between diversity and quality remains challenging and necessitates careful tuning.
Besides, even with near-optimal sampling parameters, a single LLM might still get trapped in local optima due to constraints inherent in its training data and architectural design. 

To mitigate this limitation, an alternative solution is to aggregate the specialized strengths of multiple LLMs.
Recent work~\cite{moa} has demonstrated that multiple LLMs can collaboratively enhance their instruction-following capabilities by post-editing each other's responses to the same instruction. 
Motivated by this progress, we explore leveraging the collective expertise of multiple LLMs for search-based reasoning, which, to the best of our knowledge, has not been previously tested.
Figure~\ref{fig:intro} illustrates the reasoning accuracy on the MATH-500 dataset as a function of search diversity.
The performance of search using a single LLM initially improves with increased temperature but subsequently degrades, remaining consistently lower than that of multiple-LLM search.

In this work, we propose Mixture-of-Search-Agents (\mosa{}), an advanced paradigm for step-wise search-based reasoning that aggregates the complementary strengths of multiple LLMs, leveraging both independent and collaborative contributions to search for reasoning sub-steps more effectively. 
As illustrated in Figure~\ref{fig:ours}, multiple LLMs propose diverse potential search directions at each reasoning step, either independently or through iterative refinement of each other's outputs.
This hybrid approach ensures that the reasoning process is not constrained by the limitations or biases of any single model. 
For instance, one model may excel at identifying a promising initial direction, while another might build on it to refine or extend the reasoning path. 
By combining independence and collaboration, the framework avoids local optima while enhancing reasoning accuracy in the search process.

We performed a comprehensive evaluation of \ourmethod{} across four reasoning benchmarks.
The findings indicate that \ourmethod{} consistently outperforms its single-LLM counterpart in reasoning accuracy with an average improvement of 1.71\%.
Additionally, our results indicate a synergistic interaction between multi-agent collaboration and search-based reasoning.
Further analysis and ablation studies reveal a key challenge for single-agent search-based reasoning: balancing diversity and quality varies across different benchmarks.
We also confirm a positive correlation between reasoning performance and the number of distinct search agents, validating the efficacy of multi-agent search.
Finally, experiments with an extended action set demonstrate the robustness of \ourmethod{} across diverse types of search actions.

\section{Method}
\label{sec:method}

Search-based methods have been extensively used to tackle complex reasoning tasks, such as coding and mathematics, by breaking these problems into multiple search steps~\cite{zhou2023leasttomost,yao2024tree,hao-etal-2023-reasoning}.
Our proposed paradigm is readily applicable to various search algorithms, with the Monte Carlo Tree Search (MCTS) algorithm~\cite{mcts-1,mcts-2} adopted as the search backbone in this work.
This section first introduces the baseline MCTS-based reasoning method with a single search agent~\cite{hao-etal-2023-reasoning,qi2024mutual} in $\S$~\ref{sec:method:background}, followed by our method, which leverages the expertise of multiple LLMs as search agents in $\S$~\ref{sec:method:mosa}.

\subsection{Baseline Framework}
\label{sec:method:background}

\begin{figure}[t]
\begin{center}
\centerline{\includegraphics[width=0.7\columnwidth]{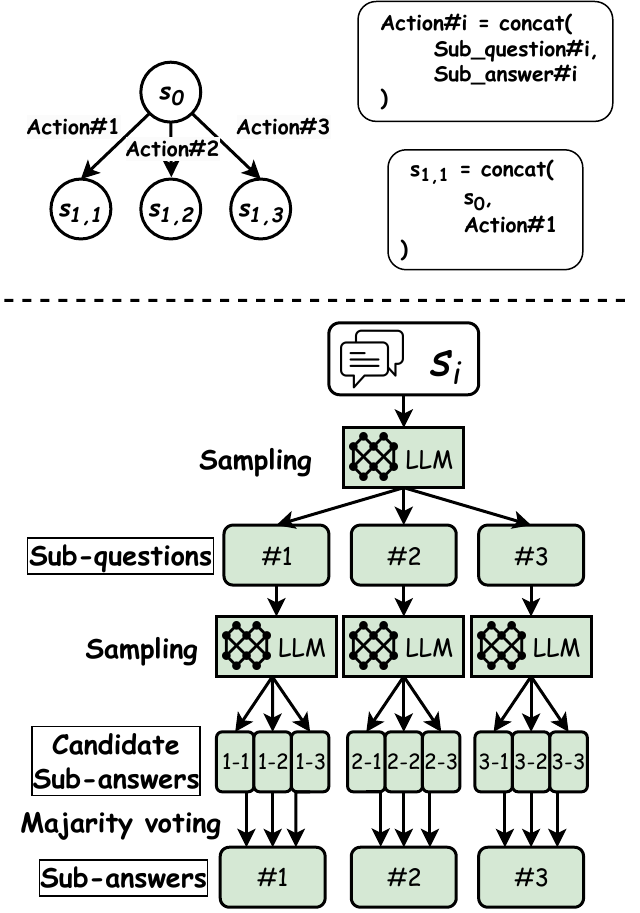}}
\caption{\emph{Top}: An overview of the root node $s_0$ and its expanded child nodes. \emph{Bottom}: The detailed framework for generating new actions (i.e., sampling sub-questions and sub-answers).}
\label{fig:baseline}
\end{center}
\vskip -0.2in
\end{figure}

\textbf{Overview}
\hspace{5pt}
Given a problem $x$ and a generator $\pi^*$, MCTS involves iteratively building a search tree starting from the root node $x$.
We first define the state space $\mathcal{S}$ and the action space $\mathcal{A}$. 
In our case, each state $s_j\in \mathcal{S}$ captures the actions (i.e., reasoning steps) generated so far alongside a specific trajectory in the search tree, while each action $a_j \in \mathcal{A}$ represents the next reasoning step based on the current state and the type of action chosen. 
As shown in the upper part of Figure~\ref{fig:baseline}, given the selected node $s_i$ (i.e., the reasoning steps generated so far), a step of \textit{Expansion} essentially creates a set of child nodes.
A child node is created by concatenating $s_i$ with the new action, with that action being the next reasoning step generated by a search agent (e.g., an LLM) given $s_i$.

\begin{algorithm*}[tb]
\begin{small}
   \caption{
   \small
   \texttt{GenerateActions}: A function for generating actions, i.e., a sub-question along with a sub-answer, given the current state.
    The implementation of this function using conventional MCTS and MoSA mainly differs in two aspects: 
   (1) In conventional MCTS methods, the number of search agents \(m=1\), while for \textsc{MoSA}, \(m > 1\). 
   (2) The \texttt{FinalizeSubAnswer} function employs heuristic majority voting for single-model search, while employing an additional neural aggregation function for \textsc{MoSA} (see the right section of Figure~\ref{fig:ours}).
   }
   
   \label{alg:mosa-A_1}
\begin{algorithmic}
    \REQUIRE Selected node \(s_i\); Number of sub-questions \(n_q\); Number of candidate sub-answers per sub-question \(n_a\); A set of LLMs \(\pi^{\mathrm{mix}} = \{ \pi_1, \pi_2, ..., \pi_m \}\)
    \ENSURE A set of new actions \(\texttt{\small new\_actions} = \{ \texttt{\small action}_1, \texttt{\small action}_2, ..., \texttt{\small action}_{n_q} \}\)
    
    \STATE Initialize \(\texttt{\small new\_actions} \leftarrow \emptyset\) \COMMENTALGO{Prepare the set of new actions}
    \FOR{\(i=1\) {\bfseries to} \(n_q\)}
        \STATE \(\pi^{\mathrm{sub\_q}} \leftarrow \texttt{\small SelectLLM}(\pi^{\mathrm{mix}})\) \COMMENTALGO{Select an LLM for generating sub-question}
        \STATE \(\texttt{\small sub\_question}_i \leftarrow \texttt{\small GenerateSubQuestion}(\pi^{\mathrm{sub\_q}}, s_i)\) \COMMENTALGO{Generate the \(i\)-th sub-question}
        
        \STATE Initialize \(\texttt{\small candidate\_sub\_answers} \leftarrow \emptyset\) \COMMENTALGO{Store candidate sub-answers for sub-question \(i\)}

        \FOR{\(j=1\) {\bfseries to} \(n_a\)}
            \STATE \(\pi^{\mathrm{sub\_a}} \leftarrow \texttt{\small SelectLLM}(\pi^{\mathrm{mix}})\) \COMMENTALGO{Select an LLM for generating a sub-answer}
            \STATE \(\texttt{\small candidate\_sub\_answer}_j \leftarrow \texttt{\small GenerateSubAnswer}(\pi^{\mathrm{sub\_a}}, s_i, \texttt{\small sub\_question}_i)\) \COMMENTALGO{Generate the \(j\)-th candidate sub-answer}
            \STATE \texttt{\small candidate\_sub\_answers.add(candidate\_sub\_answer$_j$)} \COMMENTALGO{Store the candidate sub-answer}
        \ENDFOR
        
        \STATE \(\texttt{\small sub\_answer}_i \leftarrow \texttt{\small FinalizeSubAnswer(candidate\_sub\_answers)}\) \COMMENTALGO{Aggregate or vote on candidate sub-answers}
        \STATE \(\texttt{\small action}_i \leftarrow \texttt{\small concat}(\texttt{\small sub\_question}_i, \texttt{\small sub\_answer}_i)\) \COMMENTALGO{Form the final action by concatenation}
        \STATE \texttt{\small new\_actions.add(action$_i$)} \COMMENTALGO{Add the action to the set of new actions}
    \ENDFOR
    
    \STATE return \(\texttt{\small new\_actions}\)
\end{algorithmic}
\end{small}
\end{algorithm*}

\paragraph{Action Space}
We follow rStar~\cite{qi2024mutual} to define a comprehensive set of actions into MCTS-based LLM reasoning. 
The set of actions, $A = \{A1, A2, A3, A4, A5\}$, includes:
\vspace{-10pt}
\begin{itemize}
\setlength\itemsep{0.0em}
    \item $A1$: Propose a one-step thought;
    \item $A2$: Propose the remaining thought steps;
    \item $A3$: Propose the next sub-question along with its answer;
    \item $A4$: Answer the sub-question again;
    \item $A5$: Rephrase the question.
\end{itemize}
\vspace{-10pt}
Among these actions, we designate $A3$ as the primary action, comprising a \emph{sub-question} and its corresponding \emph{sub-answer}, i.e., $\text{action}_i \equiv \text{concat}(\text{sub\_question}_i,\text{sub\_answer}_i)$.
For instance, an action can be ``\textit{\#\#\# Sub-question 3: Does the sum of the previous two digits equal 8? \#\#\# Sub-answer 3: The two digits are 3 and 5. We have $3 + 5 = 8$, so the answer is yes.}''. 
We consider the other actions along with their effects in an ablation analysis ($\S$~\ref{sec:analysis:action_set}).
We present a detailed illustration of generating new actions, i.e., combinations of sub-question \& sub-answer in Algorithm~\ref{alg:mosa-A_1}.
For a given state $s_i$,  the algorithm traverses all possible actions, where the final sub-answer for each sub-question is determined by a heuristic function, e.g., majority voting. 

\begin{figure*}[t]
\begin{center}
\centerline{\includegraphics[width=0.88\textwidth]{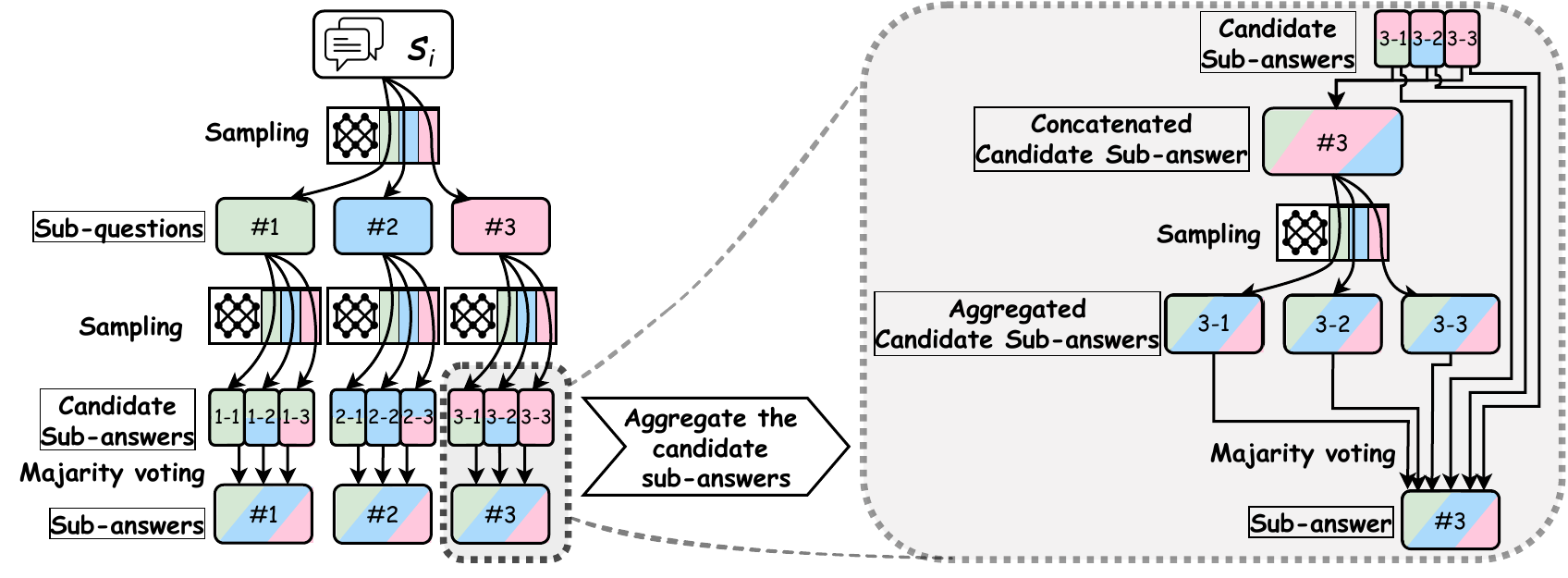}}
\caption{
Generate three new actions using \mosa. \emph{Left}: Use \mosa{} to propose sub-questions and sub-answers. \emph{Right}: Use \mosa{} to aggregate candidate sub-answers.
}
\label{fig:ours}
\end{center}
\vskip -0.2in
\end{figure*}

\textbf{Reward Function}
\hspace{5pt}
Following \citet{hao-etal-2023-reasoning,qi2024mutual}, we consider a simple yet effective reward function: actions that frequently lead to correct final answers are assigned higher rewards. 
Specifically, $Q(s,a)$, the reward value for node $s$ created by action $a$, receives a positive reward if a trajectory containing node $s$ reaches a correct final answer, and no reward otherwise. 
Since the gold answer is not available during testing, the confidence given by \emph{majority voting} is regarded as an approximation of the reward value.

\textbf{MCTS Iterations}
\hspace{5pt}
Typically, each MCTS iteration involves four steps: \textit{Selection}, \textit{Expansion}, \textit{Simulation}, and \textit{Back-propagation}.
To balance exploration and exploitation, we adopt the widely-used \textit{Upper Confidence Bounds for Trees} (UCT) algorithm~\cite{mcts-1} for \textit{Selection}.
Formally, a node $s$ is selected to maximize:
\begin{small}
\begin{equation}
    \mathrm{UCT}(s,a) = \frac{Q(s,a)}{N(s,a)} + c \sqrt{\frac{\ln{N_{\mathrm{parent}} (s)}}{N(s,a)}}
\end{equation}
\end{small}
where $N_{\mathrm{parent}} (s)$ is the number of times the parent node of $s$ has been visited, $N(s,a)$ is the number of times node $s$ has been visited, and $c$ is a constant.
Once the node $s$ is selected, an \textit{Expansion} step is performed to add child nodes to $s$.
After that, starting from a random child node, a \textit{Simulation} is performed using the default rollout policy until a terminal node is obtained or a predefined maximum depth is reached.
The outcome of the simulation determines the reward, which is then propagated back up the tree during the \textit{Back-propagation} step. 
Upon multiple iterations, we consider each leaf node as a solution. 
In this work, we focus on \textit{Expansion}, which aims to effectively expand the search space.



\textbf{Sampling Diversity}
\hspace{5pt}
Applying stochastic sampling techniques in LLM generation is essential for introducing diversity to MCTS.
As presented in the lower part of Figure~\ref{fig:baseline}, given the selected state $s_0$, the sub-questions and the sub-answer candidates are all stochastically sampled using temperature scaling, top-$k$ sampling and nucleus sampling~\cite{Holtzman2020The}.
In $\S$~\ref{sec:analysis:diversity}, we empirically alter search diversity by manipulating generation temperature for single-LLM search.


\subsection{Mixture-of-Search-Agents}
\label{sec:method:mosa}


Conventional Monte Carlo Tree Search (MCTS) methods utilizing a single model face two significant limitations:
(1) Encouraging search diversity while maintaining generation quality is challenging~\cite{tradeoff}, necessitating meticulous tuning of sampling parameters to balance the trade-off between these aspects;
(2) using heuristic metrics like majority voting to determine the final sub-answer can be less accurate when the model favors incorrect search directions.
To this end, we explore a simple yet effective alternative, Mixture-of-Search-Agents (\mosa), which employs multiple agents to perform search algorithms like MCTS and utilizes a neural function to refine the candidate step-wise outputs.
Firstly, leveraging the distinct distributions from different models intrinsically yields better generation diversity, alleviating the necessity for sampling parameters optimization.
Additionally, incorporating a neural function enhances the robustness of answer aggregation.

Figure~\ref{fig:ours} illustrates how our method generates three new actions starting from the current node $s_i$. 
Unlike vanilla single-model search, \mosa{} employs multiple agents (denoted by distinct colors) to explore diverse actions, such as sub-questions and sub-answers.
In the remainder of this section, we illustrate two roles performed by \textsc{MoSA} when generating new actions in MCTS.
Specifically, we will start with the straightforward improvement, \emph{\textsc{MoSA} as Proposers}, where multiple agents are involved for sampling actions;
then we will introduce the more intricate \emph{\textsc{MoSA} as Aggregators}, which extends the heuristic majority voting method to an aggregating phase where multiple LLMs read and refine the answers given by all.

\textbf{\textsc{MoSA} as Proposers to Diversify Actions}
\hspace{5pt}
The left side of Figure~\ref{fig:ours} shows \emph{\textsc{MoSA} as Proposers}, where the single search agent adopted by the baseline MCTS method (as in Figure~\ref{fig:baseline}) is replaced by \textsc{MoSA}. 
\textsc{MoSA} leverages multiple LLMs to enhance action diversity by fulfilling two sub-roles: multi-agent proposers that generate \emph{sub-questions} and \emph{sub-answers}.

Generating a new search action begins with sampling a sub-question from the current state $s_i$. 
The sub-question proposing phase essentially controls the directions of the current search step because whatever follows within this step is constrained by the scope of that sub-question. 
Because of this, we consider maintaining the independence among sub-questions, ensuring that the initial search direction indicated by each sub-question is independent of others. 
As shown in the upper-left part of Figure~\ref{fig:ours}, this effectively diversifies the sampled sub-questions as the same $s_i$ is colored with distinct characteristics after going through different LLMs.

After the initial search directions are created, the target is to comprehensively explore each search direction. 
To achieve this, each sub-question is answered by various LLMs, generating a diverse set of candidate sub-answers. 
These candidates are then aggregated to reach a finalized sub-answer.
A simple yet effective aggregating method is majority voting, leveraging the principle of self-consistency~\cite{wang2023selfconsistency}.



\textbf{\textsc{MoSA} as Aggregators for Collaborative Refinement}
\hspace{5pt}
We introduce a neural function, termed ``aggregator'', to mitigate the limitation of majority voting for selecting the final answer. 
An aggregator leverages the innate capability of the LLM to critique, compare and aggregate multiple answers into a final answer. 
Specifically, we prompt each LLM to consolidate all responses into an aggregated answer (see Appendix~\ref{sec:appendix:prompt} for detailed prompts), resulting in a new set of aggregated answers as illustrated in the right section of Figure~\ref{fig:ours}.
The underlying intuition is that this aggregation process enhances the likelihood of producing correct answers by facilitating comparisons among different responses, thereby increasing the overall success rates for correct answers under majority voting.
We present an example below to illustrate this intuition.

In the previous section, we consider majority-voting after obtaining candidate sub-answers from diverse proposers.
Let us consider a sub-question that requires 3 sub-answers to be generated and an \textsc{MoSA} component consisting of 3 distinct LLMs.
We simply assume that each LLM proposes one sub-answer.
If there are $k$ LLMs that are proficient at this sub-question and the other $3-k$ are not, then it is likely that we would have $k$ \emph{good} sub-answer candidates and $3-k$ \emph{bad} candidates\footnote{For clarity in illustrating our motivation, we simplify the correctness of candidate sub-answers into two groups: \emph{good} and \emph{bad}. This abstraction helps explain the role of aggregators in improving answer correctness, though actual correctness exists on a spectrum depending on task complexity and evaluation criteria.}.
With majority voting, a \emph{bad} finalized answer is likely if $k\geq2$.

Now we turn to use \textsc{MoSA} to aggregate the candidate sub-answers and then include the aggregated sub-answers into majority-voting.
The inputs for all three aggregator LLMs are the same, which concatenates the sub-question and all the three candidate sub-answers.
We hypothesize that a \emph{bad} aggregator that receives at least a \emph{good} sub-answer could yield a sub-answer that is at least better than its original \emph{bad} sub-answer.
Such a hypothesis has been empirically verified in the case of instruction following by \citet{moa}, who showed that many LLMs can generate higher-quality responses by building upon outputs from other LLMs.
Thus, if the two \emph{bad} aggregator LLMs can learn from the \emph{good} sub-answer and generate \emph{good} aggregated sub-answers, then we will have $4$ \emph{good} sub-answers and $2$ \emph{bad} ones, which lead to a \emph{good} finalized sub-answer.

\section{Experiments}
\label{sec:exp}


\subsection{Baselines}
\label{sec:exp:baselines}

\textbf{Few-shot Chain-of-Thought (CoT)}~\cite{wei2023chainofthoughtpromptingelicitsreasoning} feeds the LLM with a few demonstrations followed by the input question.
Since we are using instruction-tuned LLMs, we format the demonstrations as multi-turn dialogues.
In each turn, the human asks a question and then the assistant answers it.

\textbf{Self-Consistency@$\mathbf{\textit{n}}$}~\cite{wang2023selfconsistency} also adopts the few-shot CoT prompting scheme, but it samples $n$ independent answers per instance.
The final answer is then given by majority voting over the $n$ candidate answers.
Except for the conventional single-LLM self-consistency experiments, we also evaluate self-consistency with multiple different LLMs.
Such a multi-LLM self-consistency setting can be regarded as a simplified version of ~\citet{moa}, which collects direct answers from various agents and aggregates them with majority voting.

\textbf{Reasoning-via-Planning (RAP)}~\cite{hao-etal-2023-reasoning} is a representative LLM-based reasoning method using MCTS.
We use it as the foundation to apply \mosa.
In each search step, RAP generates one or more sub-questions along with their sub-answers.
The original RAP paper adopted different reward functions for different types of tasks.
In this work, we use the simple self-consistency score as the reward value, which has been shown to be competitive with those manually designed ones in Appendix A.1 of \citet{qi2024mutual}.
Note that we ensure the total number of LLM forward calls of a single-LLM method are approximately the same as its multi-LLM counterpart, e.g., RAP \emph{versus} RAP + \mosa{} as Proposers in Table~\ref{tab:main_results}.

\textbf{rStar}~\cite{qi2024mutual} is one of the recent SoTA MCTS-based LLM reasoning methods.
The authors proposed a comprehensive set of search actions, which we have introduced in $\S$~\ref{sec:method:background}.
We adopt their innovative set of actions to evaluate the effects brought by the scope of action set on \mosa{} in $\S$~\ref{sec:analysis:action_set}.

\begin{table*}[t]
\caption{
Main results.
Those rows marked by $\ddagger$ were reported by the rStar paper~\cite{qi2024mutual} using Llama-3-8B-Instruct. All other results are reported by our experiments.
\textsc{Multi} refers to multi-LLM while \textsc{STG} represents StrategyQA.
\textbf{The highest number} on each dataset is marked in bold while \underline{the secondary high} is underlined.
}
\label{tab:main_results}
\vskip 0.1in
\begin{center}
\begin{small}
\begin{sc}
\begin{tabular}{l|cc|cccc|cr}
\toprule
Method & Multi? & Search? & GSM8K & SVAMP & MATH & STG & Avg. \\
\midrule
Few-shot CoT:                                       &   &   &   &  &  &  & \\
\hspace{10pt} $\circ$ Llama-3.1-8B-Instruct & $\times$ & $\times$ & 84.00 & 86.80 & 41.60 & 67.39 & 69.95 \\
\hspace{10pt} $\circ$ Ministral-8B-Instruct-2410 & $\times$ & $\times$ & 82.41 & 89.20 & 40.00 & 70.60 & 70.55 \\
\hspace{10pt} $\circ$ Qwen-2-7B-Instruct & $\times$ & $\times$ & 84.00 & 88.60 & 24.20 & 66.67 & 65.87 \\
\hspace{10pt} $\circ$ GLM-4-9B-Chat & $\times$ & $\times$ & 83.85 & 89.70 & 40.00 & 71.32 & 71.22 \\
\hline
\emph{w/} Llama-3.1-8B-Instruct:                    &   &   &   &  &  &  &   \\
\hspace{10pt} $\circ$ Self-Consistency@4 & $\times$ & $\times$ & 88.02 & 89.70 & 43.80 & 69.43 & 72.74 \\
\hspace{10pt} $\circ$ Self-Consistency@32 & $\times$ & $\times$ & 90.37 & 92.40 & 44.80 & 70.89 & 74.62 \\
\hspace{10pt} $\circ$ Self-Consistency@128 & $\times$ & $\times$ & 90.98 & 93.30 & 52.20 & 71.32 & 76.95 \\
\hspace{10pt} $\circ$ Self-Consistency@256 & $\times$ & $\times$ & 90.90 & 92.90 & 53.20 & 71.03 & 77.01 \\
\hline
\emph{w/} All Four LLMs:                            &   &   &   &  &  &  &   \\
\hspace{10pt} $\circ$ Self-Consistency@4 & $\surd$ & $\times$ & 90.45 & 92.20 & 46.20 & 70.45 & 74.82 \\
\hspace{10pt} $\circ$ Self-Consistency@32 & $\surd$ & $\times$ & 90.75 & 93.20 & 52.60 & 71.76 & 77.08 \\
\hspace{10pt} $\circ$ Self-Consistency@128 & $\surd$ & $\times$ & \underline{91.21} & 93.70 & 53.80 & 72.78 & 77.87 \\
\hspace{10pt} $\circ$ Self-Consistency@256 & $\surd$ & $\times$ & 90.98 & 93.50 & 54.20 & 71.47 & 77.54 \\
\hline
RAP $\ddagger$ & $\times$ & $\surd$ & 80.59 & 85.70 & 18.80 & 68.71 & 63.45 \\
RAP & $\times$ & $\surd$ & 90.52 & 91.60 & 53.00 & 75.40 & 77.63 \\
\hspace{10pt} + Single-LLM as Aggregator & $\times$ & $\surd$ & 90.05 & 92.50 & \underline{54.80} & \underline{75.69} & 78.26 \\
\hspace{10pt} + \textsc{MoSA} as Proposers & $\surd$ & $\surd$ & 91.13 & \underline{94.50} & 54.60 & \underline{75.69} & \underline{78.98} \\
\hspace{10pt} + \mosa{} as Proposers \& Aggregators & $\surd$ & $\surd$ & \textbf{91.96} & \textbf{94.90} & \textbf{56.60} & \textbf{76.42} & \textbf{79.97} \\
\bottomrule
\end{tabular}
\end{sc}
\end{small}
\end{center}
\vskip -0.1in
\end{table*}

\subsection{Experimental Settings}
\label{sec:exp:exp_settings}

\textbf{Benchmarks}
\hspace{5pt}
We perform evaluation on four reasoning benchmarks covering different scopes, including three mathematical reasoning datasets (GSM8K~\cite{gsm8k}, SVAMP~\cite{svamp}, \mbox{MATH-500}~\cite{MATH,lightman2023lets}) and one commonsense reasoning dataset (StrategyQA~\cite{strategyqa}).

\textbf{Models}
\hspace{5pt}
We adopt four open-sourced instruction-following LLMs to formulate the LLM pool of \mosa{}: Llama-3.1-8B-Instruct~\cite{grattafiori2024llama3herdmodels}, Qwen-2-7B-Instruct~\cite{yang2024qwen2technicalreport}, Ministral-8B-Instruct-2410~\cite{ministral}, and GLM-4-9B-Chat~\cite{glm2024chatglm}.
The number of LLMs could also be made larger or smaller, depending on customized choices.
Our later experiments will show that benchmark performances are positively correlated with the number of distinct LLMs.

\textbf{Implementation Details}
\hspace{5pt}
For few-shot CoT baselines, we report the results of all four LLMs.
For other single-LLM baselines, like Self-Consistency@$n$ and RAP, we adopt \textbf{Llama-3.1-8B-Instruct} due to its competitiveness and robustness across various benchmarks.
For all experiments regarding sampling from multiple LLMs, we try to maintain a pseudo uniform distribution for the \texttt{SelectLLM} function in Algorithm~\ref{alg:mosa-A_1}.
That is, if 7 completions need to be sampled and there are 4 distinct LLMs, we manually assign each LLM to sample one completion and then uniformly sample 3 LLMs out of 4 without replacement to finish the remaining 3 completions.
Hyper-parameter settings are listed in Appendix~\ref{sec:appendix:exp_setting}.

\subsection{Main Results}
\label{sec:exp:main_results}
We report the main results on the four benchmarks in Table~\ref{tab:main_results}. Below we highlight our key findings.


\textbf{\mosa{} Leads in Reasoning Tasks}\hspace{5pt}
RAP + \mosa{} as Proposers \& Aggregators consistently yields superior performances across all datasets (GSM8K, SVAMP, \mbox{MATH-500}, StrategyQA), reaching an average performance (Avg.) of 79.97\%.
Specifically, it obtains exceptional improvements (+1.8\%) over the best baseline on the challenging MATH-500 benchmark, suggesting it is effective at handling complex reasoning problems.

\textbf{Synergistic Effect between Multi-Agent Collaboration and Search-based Reasoning}\hspace{5pt}
\mosa{} integrates two research paradigms: multi-agent collaboration and search-based reasoning. 
When applied independently, each achieves moderate improvements, but their combination yields significantly enhanced results due to synergy effects.
(\textbf{1}) Transitioning from single-agent to multi-agent:
Across all four benchmarks, transitioning from a single LLM to multiple LLMs with the best non-search baseline (Self-consistency) results in an average absolute improvement of +0.53\%. 
By contrast, transitioning from single-agent search (RAP) to multi-agent search (\mosa{} as Proposers) yields a larger average absolute improvement of +1.35\%. Augmenting with aggregators further increases the improvement from single-agent search (RAP + Single-LLM as Aggregator) to multi-agent search (\mosa{} as Proposers and Aggregators), achieving +1.71\%.
(\textbf{2}) Transitioning from non-search to search-based reasoning:
Using a single LLM, the performance gap between non-search (Self-consistency@256) and search (RAP) is +0.62\%. 
This gap widens to +1.44\% when employing multiple LLMs, showcasing the synergy between multi-agent collaboration and search. 
These results highlight that combining multi-agent collaboration with search-based reasoning yields significantly greater performance gains than applying either approach in isolation.


\textbf{Boosting Search-based Reasoning with \mosa{} as Aggregators}
\hspace{5pt}
While vanilla RAP performs well, the inclusion of aggregators, particularly with \mosa{} as Aggregators, significantly enhances performance. 
For instance, augmenting RAP with a single-LLM aggregator yields an average improvement of +0.63\%. 
This improvement increases to +0.99\% when \mosa{} as Proposers is further enhanced with \mosa{} as Aggregators.


\textbf{Search-Based Methods Excel in Complex Reasoning Tasks}
\hspace{5pt}
The best accuracy numbers on GSM8K and SVAMP, both exceeding 90\%, suggest these datasets are relatively easier. 
In contrast, MATH-500 and StrategyQA, with best scores around 55\% and 80\%, respectively, are more challenging. 
Notably, search-based methods demonstrate a clear advantage on these more complex datasets, underscoring their effectiveness in tackling intricate reasoning tasks.
Take StrategyQA as an example, the best accuracy number with non-search methods (Self-consistency) is 72.78\%, which is significantly lower than the best search counterpart (RAP) accuracy (75.69\%).

\begin{table*}[t]
\caption{
The results of \textsc{MoSA} combined with rStar~\cite{qi2024mutual}, a recent SoTA MCTS-based reasoning method that extends the set of actions.
Those results marked with $\ddagger$ were reported by the rStar paper using Llama-3-8B-Instruct. 
All other results are reported by our experiments.
The definitions of $A\{1,2,3,4,5\}$ are in $\S$~\ref{sec:method:background}.
\textbf{The highest number} on each dataset is marked in bold while \underline{the secondary high} is underlined.
}
\label{tab:with_rstar}
\vskip 0.1in
\begin{center}
\begin{small}
\begin{sc}
\begin{tabular}{l|c|cccc|cr}
\toprule
Method & Action Set & GSM8K & SVAMP & MATH & STG & Avg. \\
\midrule
RAP $\ddagger$ & $A\{2,3\}$ & 80.59 & 85.70 & 18.80 & 68.71 & 63.45 \\
RAP  & $A\{2,3\}$ &  90.52 &  91.60 &  53.00 &  75.40 & 77.63  \\
\hspace{10pt} + \textsc{MoSA} as Proposers & $A\{2,3\}$ & 91.13  & 94.50 & 54.60 & \underline{75.69} & 78.98  \\
\hspace{10pt} + \textsc{MoSA} as Proposers \& Aggregators   & $A\{2,3\}$  &  \underline{91.96} &  94.90 &  56.60 &  \textbf{76.42} & 79.97  \\
\hline
rStar $\ddagger$ & $A\{1,2,3,4,5\}$ &  88.70 & 91.89 & 38.30 & 71.47 & 72.59  \\
rStar  & $A\{1,2,3,4,5\}$ & 91.36 & 93.30 & 59.00 & 74.96 & 79.66  \\
\hspace{10pt} + \textsc{MoSA} as Proposers  & $A\{1,2,3,4,5\}$ & \underline{91.96} & \textbf{95.60} & \underline{63.20} & 75.11 & \underline{81.47} \\
\hspace{10pt} + \textsc{MoSA} as Proposers \& Aggregators & $A\{1,2,3,4,5\}$ & \textbf{92.04} & \underline{95.10} & \textbf{63.60} & 75.40 & \textbf{81.54} \\
\bottomrule
\end{tabular}
\end{sc}
\end{small}
\end{center}
\vskip -0.1in
\end{table*}


\section{Analysis}
\label{sec:exp:analysis}

We perform a comprehensive analysis on \mosa{} in this section. 
Specifically, we scale the diversity of the single-LLM search baseline in $\S$~\ref{sec:analysis:diversity} and compare it with \mosa{}.
In $\S$~\ref{sec:analysis:number_of_agents}, we vary the number of distinct LLMs in \mosa{}.
In $\S$~\ref{sec:analysis:action_set}, we combine \mosa{} with the rich set of actions proposed by \citet{qi2024mutual}.
Finally, we evaluate variations of \mosa{} by ablating the numbers of proposers and aggregators in $\S$~\ref{sec:analysis:number_of_pro_agg}.

\subsection{Diversity \textit{versus} Performance}

\label{sec:analysis:diversity}

\begin{figure}[t]
\vskip 0.2in
\begin{center}

\includegraphics[width=0.84\columnwidth]{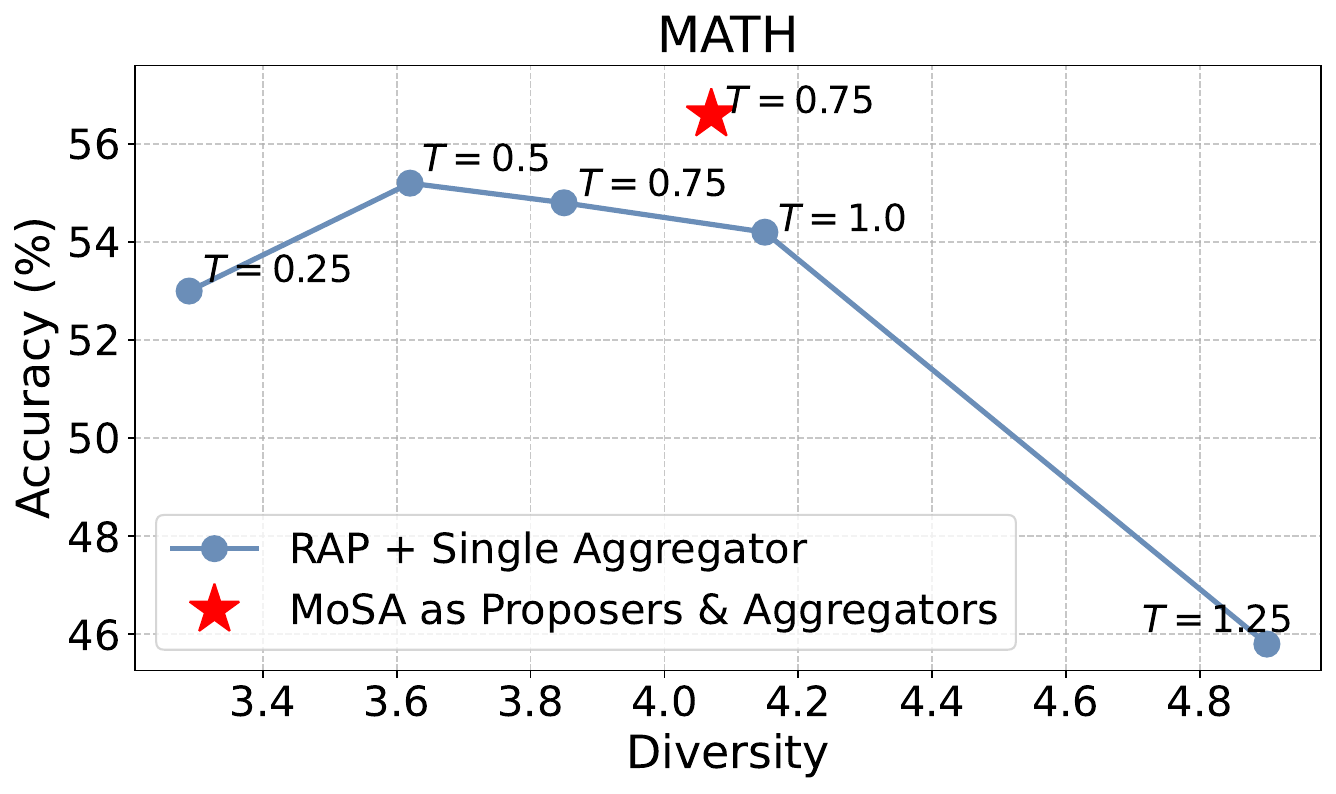}

\includegraphics[width=0.84\columnwidth]{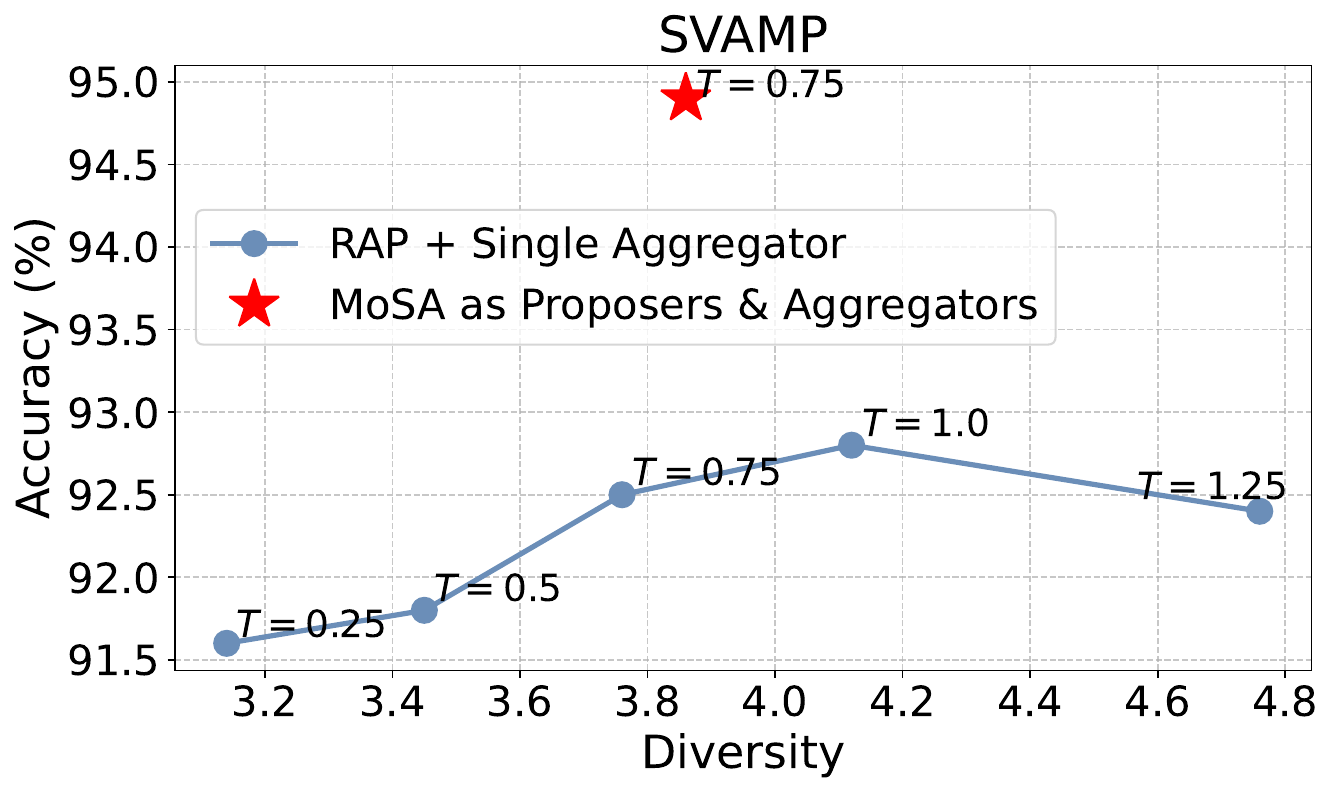}

\caption{
Diversity \emph{versus} accuracy. \textit{T} = Temperature.
}
\label{fig:diveristy_vs_acc}
\end{center}
\vskip -0.3in
\end{figure}

For single-LLM search, a common technique to increase generation diversity is to manipulate with decoding hyper-parameters, e.g., the sampling temperature. 
We modify the temperature of the RAP + Single-LLM as Aggregator baseline on two datasets, with $T=\{0.25, 0.5, 0.75, 1.0, 1.25\}$. 
Diversity is assessed by calculating the $\{1,2,3,4\}$-gram Vendi Score~\cite{friedman2023vendi} across search trajectories.
Figure~\ref{fig:diveristy_vs_acc} illustrates that while the reasoning accuracy of RAP initially improves with increasing diversity, it subsequently declines. 
More importantly, the two benchmarks favor different temperature values.
This suggests that attaining an optimal balance between diversity and reasoning performance requires careful tuning, as balancing diversity and quality can be challenging~\cite{tradeoff}.
In contrast, RAP + \mosa{} with the default sampling parameters consistently holds an advantageous position.



\subsection{Ablation of LLM Collaboration}
\label{sec:analysis:number_of_agents}

\begin{figure}[t]
    \vskip 0.2in
    \centering
    \begin{subfigure}
        \centering
        \includegraphics[width=0.23\textwidth]{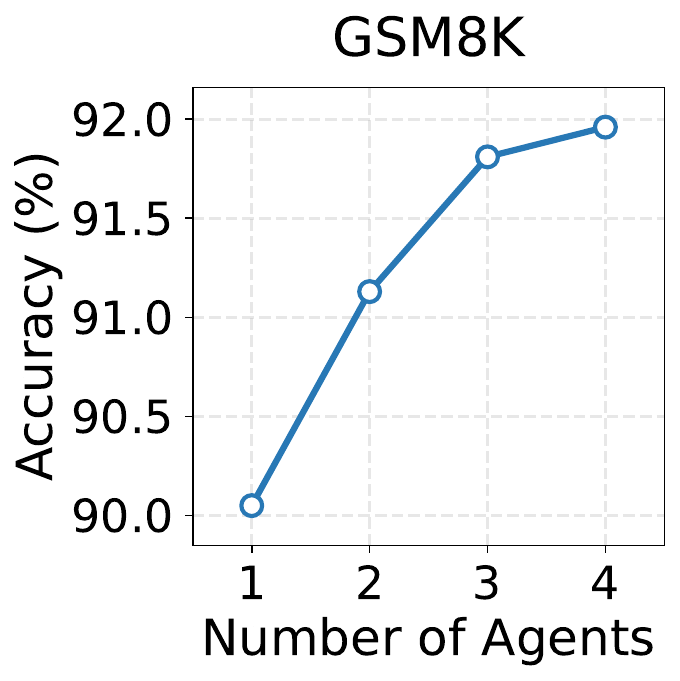}
    \end{subfigure}%
    \hfill
    \begin{subfigure}
        \centering
        \includegraphics[width=0.23\textwidth]{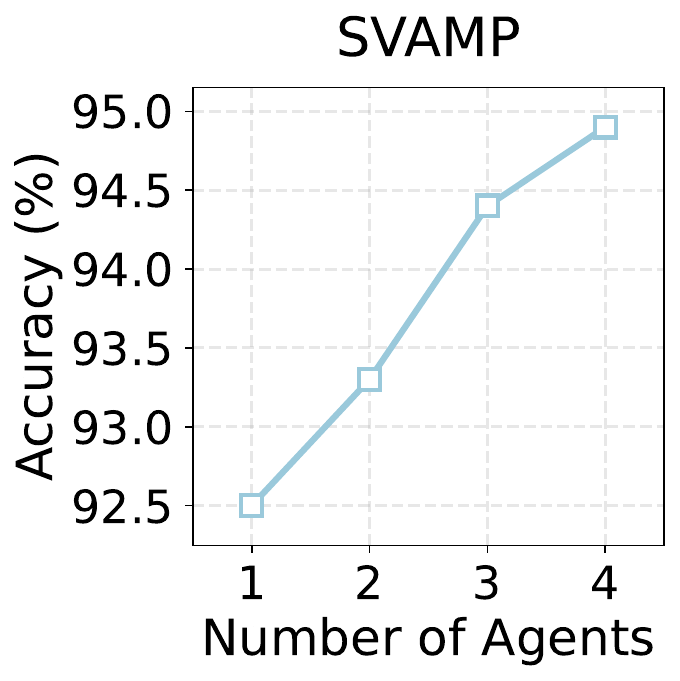}
    \end{subfigure}%
    \\
    \begin{subfigure}
        \centering
        \includegraphics[width=0.23\textwidth]{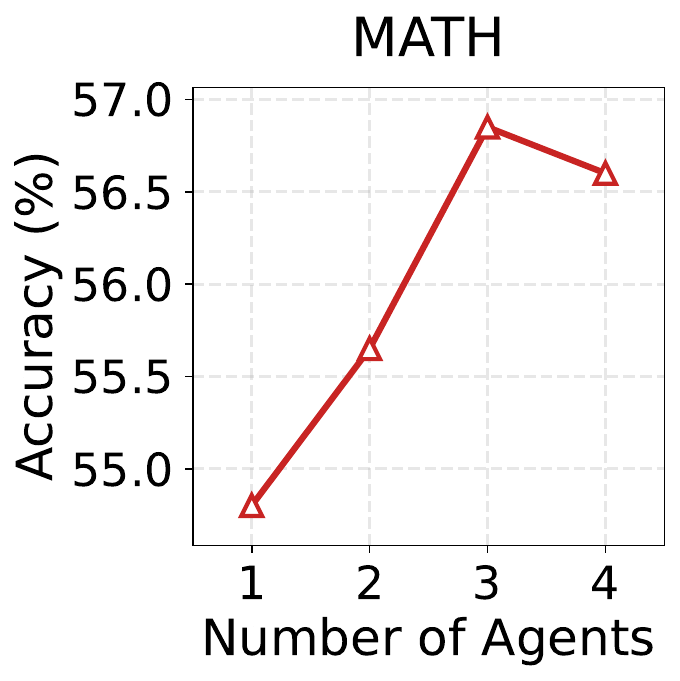}
    \end{subfigure}%
    \hfill
    \begin{subfigure}
        \centering
        \includegraphics[width=0.23\textwidth]{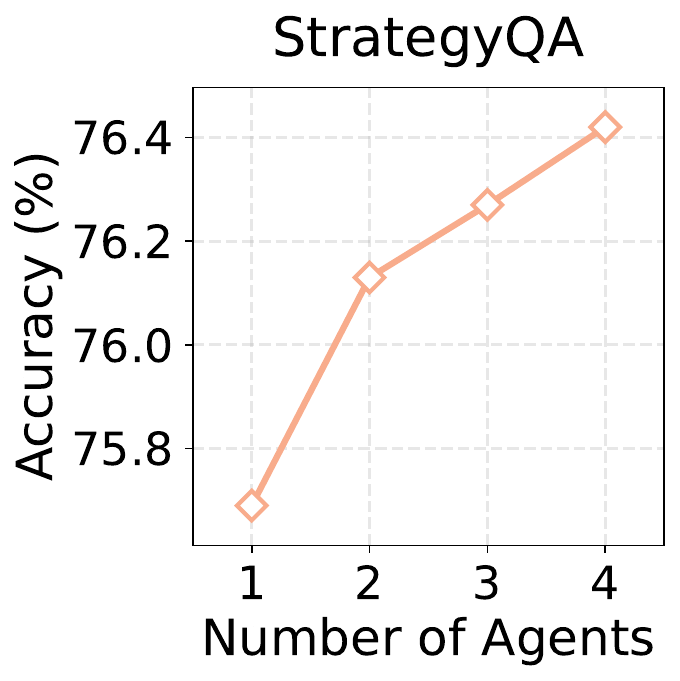}
    \end{subfigure}%
    \vskip -0.1in
    \caption{
    Reasoning accuracy with different number of distinct LLMs as search agents.
    }
    \label{fig:num_of_agents}
    \vskip -0.2in
\end{figure}

To evaluate the impact of varying the number of different LLMs in \mosa{}, we conduct an analysis using 1 to 4 LLMs across four benchmarks, prioritizing them in the following order: Llama, GLM, Qwen, Ministral.
All four variants utilize approximately the same number of LLM forward calls, ensuring that the only variable is the number of distinct LLMs involved. 
Figure~\ref{fig:num_of_agents} shows that increasing the number of different LLMs generally correlates with higher reasoning accuracy, except for a slight decrease in performance when the number of LLMs increases from 3 to 4 on MATH-500. 
This trend indicates that the diverse expertise contributed by different LLMs significantly enhances search-based reasoning performance.





\subsection{Support for Extended Action Set}
\label{sec:analysis:action_set}

rStar~\cite{qi2024mutual} proposes using a comprehensive set of actions in MCTS-based LLM reasoning. 
Since enriching the action set is orthogonal to our method, we hypothesize that \mosa{} is compatible with the enlarged action set. 
The results in Table~\ref{tab:with_rstar} support our hypothesis. 
For example, rStar combined with \ourmethod{} boosts the reasoning accuracy on MATH-500 from 59.00\% to 63.20\% (+ \mosa{} as Proposers) and 63.60\% (+ \mosa{} as Proposers \& Aggregators). 
We also found that enriching the action set is not always beneficial. 
On StrategyQA, the expanded action set yielded inferior performance; however, we note that \mosa{} still demonstrates improvements.

\begin{table*}[t]
\caption{
Ablation analysis to isolate the effects of \mosa{} as proposers and as aggregators, respectively, for search-based reasoning.
By multi, we are referring to the default setting in our experiments, i.e., 4 distinct LLMs.
\textbf{The highest number} on each dataset is marked in bold while \underline{the secondary high} is underlined.
}
\label{tab:ablation}
\vskip 0.15in
\begin{center}
\begin{small}
\begin{sc}
\begin{tabular}{cc|cccc|cr}
\toprule
\#Proposer(s) & \#Aggregator(s) & GSM8K & SVAMP & MATH & StrategyQA  & Avg. \\
\midrule
Single & None & 90.52 & 91.60 & 53.00 & 75.40 & 77.63 \\
Single & Single & 90.05  & 92.50 & 54.80 & 75.69 & 78.26  \\
Single & Multi  & 91.05  & 91.90 & 55.60 & \underline{76.42} & 78.74 \\
Multi & None  & 91.13  & \underline{94.50} & 54.60 & 75.69 & 78.98 \\
Multi & Single  &  \underline{91.66} & 94.20 & \underline{56.00} & 76.13 & \underline{79.50} \\
Multi & Multi  &  \textbf{91.96} &  \textbf{94.90} &  \textbf{56.60} &  \textbf{76.42} & \textbf{79.97} \\
\bottomrule
\end{tabular}
\end{sc}
\end{small}
\end{center}
\vskip -0.1in
\end{table*}

\subsection{Ablation of Proposers \& Aggregators}
\label{sec:analysis:number_of_pro_agg}

    



    

We consider to isolate the effects of \mosa{} as Proposers and \mosa{} as Aggregators by ablating the number of distinct LLMs for those two roles.
As shown in Table~\ref{tab:ablation}, changing the number of distinct proposers to be single yields a larger decrease comparing with ablating the number of aggregators (-1.23\% \emph{versus} -0.47\%), suggesting that \mosa{} brings more benefits as proposers.

\section{Related Work}

\subsection{Reasoning with LLMs}
The recent focus on large language models is partly due to their exceptional performance in solving complex reasoning tasks.
A prominent example is Chain-of-Thought (CoT) reasoning~\cite{wei2023chainofthoughtpromptingelicitsreasoning}.
Recent advancements include self-consistency~\cite{wang2023selfconsistency}, problem decomposition~\cite{zhou2023leasttomost}, the use of tools~\cite{pal,chen2023program}, and search-based methods~\cite{hao-etal-2023-reasoning,yao2024tree,qi2024mutual}. 
Among these approaches, \mosa{} is most closely aligned with search-based reasoning methods.

\textbf{Search-based Reasoning}
\hspace{5pt}
Search-based reasoning has demonstrated effectiveness, particularly for solving complex, multi-step problems~\cite{hao-etal-2023-reasoning,yao2024tree,chen2024understandingtreethoughtssucceeds,zhang2024accessinggpt4levelmathematical,chen2024alphamathzeroprocesssupervision,qi2024mutual,zhang2024restmctsllmselftrainingprocess,zhou2023language,koh2024treesearchlanguagemodel}. 
One of the recent state-of-the-art systems in this domain is rStar~\cite{qi2024mutual}. 
rStar introduces two key innovations: (1) expanding the Monte Carlo Tree Search (MCTS) action space from one or two actions to five; and (2) employing a secondary LLM to verify the reasoning trajectories generated by the primary LLM through MCTS.
In $\S$~\ref{sec:analysis:action_set}, we empirically demonstrate that our method is complementary to the enriched action set of rStar.

\subsection{LLM Ensemble}
Ensembling, a widely used technique for leveraging the strengths of multiple models, remains highly effective in the era of LLMs.
\citet{jiang-etal-2023-llm} proposed pairwise reranking of LLM outputs and fusing multiple responses using a trained generative model. 
Several studies have proposed training routing functions to match queries with appropriate LLMs~\cite{lu2023routingexpertefficientrewardguided,shnitzer2023largelanguagemodelrouting,wang2024fusing}. 
Others have proposed averaging the output distributions of multiple LLMs~\cite{huang2024ensemblelearningheterogeneouslarge}.

Another line of research focuses on multi-agent collaboration, where multiple LLMs interact to discuss or debate specific topics~\cite{du2023improvingfactualityreasoninglanguage,liang-etal-2024-encouraging,chan2023chatevalbetterllmbasedevaluators,xu2023reasoninglargelanguagemodels,liu2024a,he-etal-2023-lego,chen-etal-2024-reconcile,zhang-etal-2024-exploring}. 
Common design variations in this paradigm include role assignments for LLMs (e.g., debaters and judges) and discussion mechanisms (e.g., symmetric versus asymmetric interactions).


\section{Conclusion}
\label{sec:conclusion}
In this work, we investigated a novel paradigm called \mosa{}.
\mosa{} combines independent exploration and iterative refinement among multiple LLMs to enhance reasoning diversity and accuracy. 
Experiments across benchmarks demonstrate its consistent advantages over single-LLM and multi-agent baselines, especially in complex tasks. 
This work underscores the potential of multi-agent collaboration in advancing search-based reasoning.


\section*{Impact Statement}
This work aims to contribute to the advancement of reasoning with LLMs. 
While our research could have various societal implications, none are deemed significant enough to warrant specific mention at this stage.




\bibliography{main}
\bibliographystyle{icml2025}

\newpage
\appendix



\onecolumn

\section{Additional Experimental Settings}
\label{sec:appendix:exp_setting}

\subsection{Hyper-parameters}
The default sampling parameters for LLM generation are \{temperature=0.75, top\_k=40, top\_p=0.95\}.
Across all MCTS experiments, we set the number of rollouts to 8, the number of sub-questions per node to 4, the number of candidate sub-answers per sub-question to 4, the maximum depth allowed to 5.

\subsection{Dataset Statistics}
Since we make use of the rStar code base~\footnote{\url{https://github.com/zhentingqi/rStar/}} to implement \mosa, we directly adopt the data files released in their git repository.
There are 1,319 instances in GSM8K, 1,000 instances in SVAMP, 500 instances in MATH-500, and 687 instances in StrategyQA.

\section{Additional Implementation Details for Aggregators}
\label{sec:appendix:prompt}

In this section, we will show the basic instruction and several in-context learning demonstrations for aggregators.

\newpage

\begin{center}
	\small
	\begin{tcolorbox}[width=1\linewidth,title={\textbf{Basic Instruction for Aggregators}}]
		**TASK**:
        
        You are an intelligent and supportive AI assistant. You will receive a collection of responses from various AI assistants regarding a query. Your goal is to synthesize these responses into a single, high-quality response. You should first write down your thoughts as "\#\#\# Thoughts: \textless{}THOUGHTS\textgreater{}.", and then write down the synthesized response in the format of "\#\#\# Synthesized Response: \textless{}SYNTHESIZED RESPONSE\textgreater{}.". If "The answer is: \textless{}ANSWER\textgreater{}" appears in the set of responses, you should also write down "The answer is: \textless{}ANSWER\textgreater{}"; otherwise, do not write it. 
        
        1. Critically assess the information in the responses, as some may be biased, incorrect, or nonsensical. 
        
        2. Your synthesized response should not merely echo the provided answers but should present a refined, accurate, and thorough response to the prompt. 
        
        3.Ensure that your response is well-organized, coherent, and meets the highest standards of accuracy and reliability.
	\end{tcolorbox}
\end{center}

\newpage

\begin{center}
	\small
	\begin{tcolorbox}[width=1\linewidth,title={\textbf{An Example ICL Demonstration of MATH-500 for Aggregators: User Query}}]

\textbf{User:}

\textbf{[Here is the user query that was fed into those AI assistants]:}

Given a question, please decompose it into sub-questions. For each sub-question, please answer it in a complete sentence, ending with "The answer is \textless a numeric answer \textgreater". When the original question is answerable, please start the subquestion with "Now we can answer the question: \textless{}original question\textgreater{}".

Question 5: Find the product $CD$ of the integers $C$ and $D$ for which

\[\frac{C}{x-3}+\frac{D}{x+8}=\frac{4x-23}{x^2+5x-24}\]for all real values of $x$ except $-8$ and $3$.

Question 5.1: What is the equation obtained by multiplying both sides of the given equation by $(x - 3)(x + 8)$?

Answer 5.1: $(x - 3)(x + 8) \cdot \frac{C}{x-3} + (x - 3)(x + 8) \cdot \frac{D}{x+8} = (x - 3)(x + 8) \cdot \frac{4x-23}{x^2+5x-24}$

This simplifies to $C(x + 8) + D(x - 3) = 4x - 23$.

The answer is $C(x + 8) + D(x - 3) = 4x - 23$.

Question 5.2: What are the values of $C$ and $D$ that satisfy the equation $C(x + 8) + D(x - 3) = 4x - 23$ for all $x$ except $-8$ and $3$?

Answer 5.2:

\textbf{[Here are the responses from various AI assistants]:}

Response 1:

To find the values of $C$ and $D$, we can substitute specific values of $x$ into the equation $C(x + 8) + D(x - 3) = 4x - 23$ to obtain a system of equations. 

Question 5.2.1: What happens if we substitute $x = 3$ into the equation $C(x + 8) + D(x - 3) = 4x - 23$?

Answer 5.2.1: Substituting $x = 3$ gives $C(3 + 8) + D(3 - 3) = 4 \cdot 3 - 23$, which simplifies to $11C = -11$. The answer is $11C = -11$.

Question 5.2.2: What happens if we substitute $x = -8$ into the equation $C(x + 8) + D(x - 3) = 4x - 23$?

Answer 5.2.2: Substituting $x = -8$ gives $C(-8 + 8) + D(-8 - 3) = 4 \cdot (-8) - 23$, which simplifies to $-11D = -59$. The answer is $-11D = -59$.

Question 5.2.3: Now we can answer the question: What are the values of $C$ and $D$?

Answer 5.2.3: Solving the system of equations $11C = -11$ and $-11D = -59$ gives $C = -1$ and $D = 7/11 \cdot 11 = 7$. The answer is -1 and 7 respectively.

Response 2:

To find the values of $C$ and $D$, we can compare the coefficients of the corresponding terms on both sides of the equation.

For the $x^2$ terms, we have $C + D = 0$.

For the $x$ terms, we have $8C - 3D = 4$.

Solving these equations simultaneously, we find that $C = 3$ and $D = -3$.

Therefore, the product $CD = 3 \cdot (-3) = \boxed{-9}$.

Response 3:

To find $C$ and $D$, we can substitute $x = -8$ and $x = 3$ into the equation $C(x + 8) + D(x - 3) = 4x - 23$.

For $x = -8$:
\[C(-8 + 8) + D(-8 - 3) = 4(-8) - 23\]
\[-11D = -55\]
\[D = 5\]

For $x = 3$:
\[C(3 + 8) + D(3 - 3) = 4(3) - 23\]
\[11C = -13\]
\[C = -\frac{13}{11}\]

Now we can answer the question: What is the product $CD$ of the integers $C$ and $D$?
\[CD = \left(-\frac{13}{11}\right)(5) = -\frac{65}{11}\]
The answer is $-\frac{65}{11}$.

Response 4:

To find $C$ and $D$, we can set $x = 3$ and $x = -8$ to get two equations:

1. $C(3 + 8) + D(3 - 3) = 4(3) - 23 \Rightarrow 11C = -11 \Rightarrow C = -1$

2. $C(-8 + 8) + D(-8 - 3) = 4(-8) - 23 \Rightarrow -11D = -55 \Rightarrow D = 5$

The answer is $C = -1$ and $D = 5$.

	\end{tcolorbox}
\end{center}

\newpage

\begin{center}
	\small
	\begin{tcolorbox}[width=1\linewidth,title={\textbf{An Example ICL Demonstration of MATH-500 for Aggregators: Assistant Response}}]

\textbf{Assistant:}

\#\#\# Thoughts:

1. The responses generally attempt to solve the equation \(C(x + 8) + D(x - 3) = 4x - 23\) by substituting specific values for \(x\), namely \(x = 3\) and \(x = -8\).

2. The correct method involves creating a system of linear equations by substituting these values and solving for \(C\) and \(D\).

3. There is a discrepancy in the values of \(C\) and \(D\) obtained in the responses:

   - Some responses correctly find \(C = -1\) and \(D = 5\).
   
   - Another response incorrectly finds \(C = -13/11\) and \(D = 5\).
   
   - Another response finds \(C = 3\) and \(D = -3\), which is not consistent with the given equation.

4. The correct values should satisfy both equations derived from the substitutions.

\#\#\# Synthesized Response:
 
 The equation \(C(x + 8) + D(x - 3) = 4x - 23\) can be solved by substituting specific values for \(x\):

1. Substituting \(x = 3\):
\[C(3 + 8) + D(3 - 3) = 4(3) - 23\]
\[11C = -11 \]
\[C = -1\]

2. Substituting \(x = -8\):
\[C(-8 + 8) + D(-8 - 3) = 4(-8) - 23\]
\[-11D = -55 \]
\[D = 5\]

Thus, the values of \(C\) and \(D\) are \(C = -1\) and \(D = 5\). The product \(CD\) is:
\[CD = (-1) \cdot 5 = -5\]

The answer is: -5

	\end{tcolorbox}
\end{center}

\newpage

\begin{center}
	\small
	\begin{tcolorbox}[width=1\linewidth,title={\textbf{An Example ICL Demonstration of StrategyQA for Aggregators: User Query}}]

\textbf{User:}

\textbf{[Here is the user query that was fed into those AI assistants]:}

Given a question, please decompose it into sub-questions. For each sub-question, please answer it in a complete sentence. When the original question is answerable, please start the sub-question with "Now we can answer the question: \textless{}original question\textgreater{}".

Question 5: Johnny Carson: American talk show host and comedian. Could Johnny Carson's children fill out a water polo team?

Question 5.1: How many children did Johnny Carson have?

Answer 5.1: Johnny Carson had three children: two daughters, Catherine and Christie, and a son, Jeff.

Question 5.2: How many players are needed to form a water polo team?

Answer 5.2:

\textbf{[Here are the responses from various AI assistants]:}

Response 1:

A standard water polo team consists of 13 players.

Response 2:

A water polo team consists of seven players, including one goalkeeper and six field players.

Response 3:

A water polo team consists of 13 players: 1 goalkeeper and 12 field players (7 field players and 5 substitutes).

Response 4:

A standard water polo team consists of 13 players, including 7 field players and 6 substitutes.

	\end{tcolorbox}
\end{center}

\newpage

\begin{center}
	\small
	\begin{tcolorbox}[width=1\linewidth,title={\textbf{An Example ICL Demonstration of StrategyQA for Aggregators: Assistant Response}}]

\textbf{Assistant:}

\#\#\# Thoughts:

1. The responses contain valuable information about the composition of a water polo team. 

2. There is a consensus that a water polo team consists of 13 players in total. However, there is some variation in how these 13 players are categorized.

3. The key point to clarify is how many players are in the water at one time and how many are substitutes.

4. It appears that in the water during play, there are 7 players (including the goalkeeper), and the remaining players are substitutes.

\#\#\# Synthesized Response:

The synthesized response is: 

To form a water polo team, a total of 13 players are required. This team includes 7 players who are in the water during play, comprising one goalkeeper and six field players. The remaining 6 players serve as substitutes.

	\end{tcolorbox}
\end{center}


\end{document}